%% file: main.tex
\documentclass[10pt,twocolumn,letterpaper]{article}

\usepackage{wacv}              

\usepackage[accsupp]{axessibility} 
\usepackage{amsmath}
\usepackage{amssymb}
\usepackage{booktabs}
\usepackage{threeparttable}
\usepackage{multirow}
\usepackage{subcaption}
\usepackage{bm}
\usepackage{cite}
\usepackage{multirow}
\usepackage{booktabs}
\usepackage{url}
\usepackage{caption}
\usepackage{subcaption}
\usepackage{textcomp}
\usepackage{cite}
\usepackage{amsmath,amssymb,amsfonts}
\usepackage{algorithmic}
\usepackage[linesnumbered,ruled,vlined]{algorithm2e}
\usepackage{float}
\usepackage{graphicx}
\usepackage{textcomp}
\usepackage{xcolor}
\usepackage{afterpage}
\usepackage{comment}
\usepackage{blindtext}
\usepackage{siunitx}
\usepackage{balance}
\usepackage{bm}

\usepackage[utf8]{inputenc}
\usepackage{multirow}
\usepackage{booktabs}
\usepackage{url}

\usepackage{pifont}
\usepackage[linesnumbered,ruled,vlined]{algorithm2e}

\usepackage[pagebackref,breaklinks,colorlinks]{hyperref}

\usepackage[capitalize]{cleveref}
\crefname{section}{Sec.}{Secs.}
\Crefname{section}{Section}{Sections}
\Crefname{table}{Table}{Tables}
\crefname{table}{Tab.}{Tabs.}

\def\wacvPaperID{2295} 
\def\confName{WACV}
\def\confYear{2025}

\begin{document}

\title{Uncertainty Awareness Enables Efficient Labeling for Cancer Subtyping in Digital Pathology}



\author{
    Nirhoshan Sivaroopan$^{1,\dagger}$ \and 
    Chamuditha Jayanga Galappaththige$^1$ \and 
    Chalani Ekanayake$^1$ \and 
    Hasindri Watawana$^1$ \and 
    Ranga Rodrigo$^1$ \and 
    Chamira U. S. Edussooriya$^{1}$ \and 
    Dushan N. Wadduwage$^{2,3,\ddagger,}$\thanks{Corresponding author} \\
    {\small $^1$University of Moratuwa \quad $^2$Harvard University \quad $^3$Old Dominion University} \\
    \vspace{3mm}
    {\small \texttt{$^\dagger$180428t@uom.lk, $^\ddagger$dwadduwa@odu.edu}} 
}

\maketitle

\input{Abstract}

%

\input{Introduction}
\input{Methodology}
\input{Results}

\input{Conclusion}



\newpage
{\small
\bibliographystyle{ieee_fullname}
\bibliography{sample}
}

\end{document}

%% file: Abstract.tex
\begin{abstract}
Machine-learning-assisted cancer subtyping is a promising avenue in digital pathology. Cancer subtyping models however require careful training using expert annotations, so that they can be inferred with a degree of known certainty (or uncertainty).  To this end, we introduce the concept of uncertainty awareness into a self-supervised contrastive learning model. This is achieved by computing an evidence vector at every epoch, which assesses the model’s confidence in its predictions. The derived uncertainty score is then utilized as a metric to selectively label the most crucial images that require further annotation, thus iteratively refining the training process. With just 1-10\% of strategically selected annotations, we attain state-of-the-art performance in cancer subtyping on benchmark datasets. Our method not only strategically guides the annotation process to minimize the need for extensive labeled datasets, but also improve the precision and efficiency of classifications. This development is particularly beneficial in settings where the availability of labeled data is limited, offering a promising direction for future research and application in digital pathology. Our code is available at \href{https://github.com/Nirhoshan/AI-for-histopathology}{https://github.com/Nirhoshan/AI-for-histopathology}

\end{abstract}

%% file: Introduction.tex
\section{Introduction}
\label{sec:intro}

Integration of deep learning into computer-assisted digital pathology has revolutionized cancer diagnostics, offering a powerful tool to streamline the complex, labor-intensive, and error-prone processes associated with image-based detection. Despite these advancements, the field faces a significant challenge: the exhaustive and costly process of image annotation. Histopathological analysis demands precise labeling by expert pathologists, a procedure that is not only time-consuming but also heavily resource-dependent. Addressing this issue, the community initially turned to self-supervised learning (SSL) as a solution \cite{SelfSupervisedHisto, luo2022self}, which, while effective in some respects, often lacked in providing explainable model predictions, a critical requirement for potential clinical use. These models, focused mainly on accuracy, frequently underperformed on datasets with limited domain similarity to the training data \cite{seth2023fusdom}.

Another approach to mitigate the annotation burden is active learning (AL). AL introduces a human-in-the-loop querying strategy, offering a degree of interpretability and reduction of labeling effort \cite{settles2011theories}. Research such as that conducted by \cite{carse2019active} investigated different querying strategies in AL, finding that random sampling frequently surpasses strategic label selection in patch-based machine learning. This observation led to the adoption of a method where multiple patches are combined for AL, albeit at the expense of higher computational requirements. 

A key enhancement to AL lies in incorporating uncertainty into the querying process~\cite{jin2021reducing}, as it directly influences the explainability and reliability~\cite{hang2022reliability} of model predictions.

Previous works, such as~\cite{jin2021reducing}, have employed uncertainty metrics at the Whole Slide Image (WSI) level. However, these approaches often lacked interpretability, as they relied  on indirect uncertainty parameters such as dropouts or model weights.

Our work synergizes SSL and AL, addressing the shortcomings of each method when used individually. We begin by evaluating various models across different labeling scenarios to identify the most suitable SSL framework. This led us to select SimCLRv2 \cite{chen2020big} for its exceptional performance. We then enhance this framework by integrating a novel strategy of modeling uncertainty within the architecture itself \cite{sensoy2018evidential}, an approach not previously employed in histopathology. This addition not only boosts the interpretability of the model predictions---a vital aspect in critical domains like histopathology---but also sets a new standard in the field.

Leveraging this enhanced SSL framework, we then apply the uncertainty score as a querying strategy in AL. This approach is benchmarked against traditional random sampling methods. Our combined SSL and AL framework excels in patch-level classification for binary and multi-class cancer types. It adds explainability to model predictions and significantly reduces annotation efforts. The results are compelling: our model achieves parity with state-of-the-art (SOTA) outcomes using only 2-3\% of labels and surpasses them at the 9\% label mark. The subsequent sections of this paper detail our process in achieving these results, from the selection and enhancement of the SSL framework to the application of uncertainty-aware querying in AL for patch level classification.

 \section{Related Work}

\subsection{Self-supervised Representation Learning}
SSL has emerged as a powerful approach, especially for pre-training large models using unlabeled data \cite{Misra2019SelfSupervisedLO,what,byol,sim,obow,barlow}. In the realm of digital pathology, SSL frameworks like SimCLR \cite{chen2020simple} and its enhanced version, SimCLRv2 \cite{chen2020big}, have shown promise by learning rich representations from relatively large amount of unlabeled data. SimCLRv2, in particular, improves upon its predecessor through larger backbone networks, an expanded projection head, and the application of knowledge distillation \cite{hinton2015distilling}. Another notable SSL method, Masked Auto-Encoding (MAE) \cite{mae}, reconstructs images from partially masked inputs, demonstrating its utility in tasks like image classification. While these SSL methods, including adaptations for digital pathology like those by \cite{SelfSupervisedHisto} and \cite{luo2022self}, have proven effective, SimCLRv2's adoption in digital pathology remains unexplored. Moreover, the potential of these models, particularly in terms of the uncertainty in their predictions, has yet to be fully investigated.


\subsection{Uncertainty Quantification}
Estimating uncertainty in deep learning models is a crucial yet challenging aspect of machine learning, particularly in clinical applications like histopathology. Traditional methods, such as Monte Carlo dropout \cite{wang2013fast,gal2016dropout}, deep ensembles \cite{lakshminarayanan2017simple,wenzel2020hyperparameter}, and test-time augmentation \cite{dolezal2022uncertainty}, generate variability in predictions have been used to estimate uncertainty. However, these approaches often rely on inherent ambiguities in model parameters, lacking precise mathematical quantification of uncertainty. Recognizing this limitation, our work enhances SSL framework with a Bayesian approach to uncertainty estimation, as proposed by \cite{sensoy2018evidential}. This approach, grounded in the theory of evidence, excels in task-agnostic learning across different domains and it aligns perfectly with our objective of harnessing public datasets in digital pathology to acquire extensive pre-trained domain knowledge.

In our exploration of uncertainty, it is important to acknowledge the two primary types: aleatoric and epistemic uncertainty. Aleatoric uncertainty refers to the inherent noise in the data that cannot be reduced through model training such as image quality, while epistemic uncertainty is related to the model's lack of knowledge, which can be mitigated through better training and data representation. In this work, we focus on reducing epistemic uncertainty by quantifying the aspects of uncertainty that are within our control through careful training strategies. By addressing epistemic uncertainty, we aim to enhance model reliability and improve performance in clinical applications.

\subsection{Active Learning}
\label{subsec:al}

AL is the machine learning method which actively queries the most informative labels to consistently improve the model training. AL is a well adapted method in histopathology to reduce annotation cost ~\cite{jin2021reducing}, ~\cite{carse2019active} . In AL the network training will be initiated  by labelling a limited number of randomly selected images. Then, the key problem in AL is how the querying strategy is defined to select the most valuable samples to gather the most information for model training. Researches conducted for AL were circling around the challenge of identifying the most effective querying strategy to find images that yield the highest entropy. In ~\cite{carse2019active}, the researchers compared random sampling with different querying strategies. In ~\cite{jin2021reducing}, authors turned into quantifying uncertainty to find the images with highest entropy. However, in ~\cite{du2018breast}, authors tried to improve the AL by incorporating both samples with high entropy values and low entropy values to emphasize the confidence boosting. Though all these work introduced different querying strategies to reduce the annotation cost, the accuracy was on par or slightly less~\cite{du2018breast} compared to that with random selection of images. This may be due to the querying strategy failing to select the most informative images for the next iteration. We leverage the uncertainty estimation method introduced in ~\cite{sensoy2018evidential}, that proved to be performing better in uncertainty quantification compared to other uncertainty estimating methods ~\cite{sensoy2018evidential}, to develop the querying strategy, which resulted in reduced annotation cost and significant improvement in accuracy compared to the random sampling of labels.

%% file: Methodology.tex
\section{Methodology}
\label{sec:Methodology}

\begin{figure*}[htbp]
    \centering
    \includegraphics[width=\linewidth]{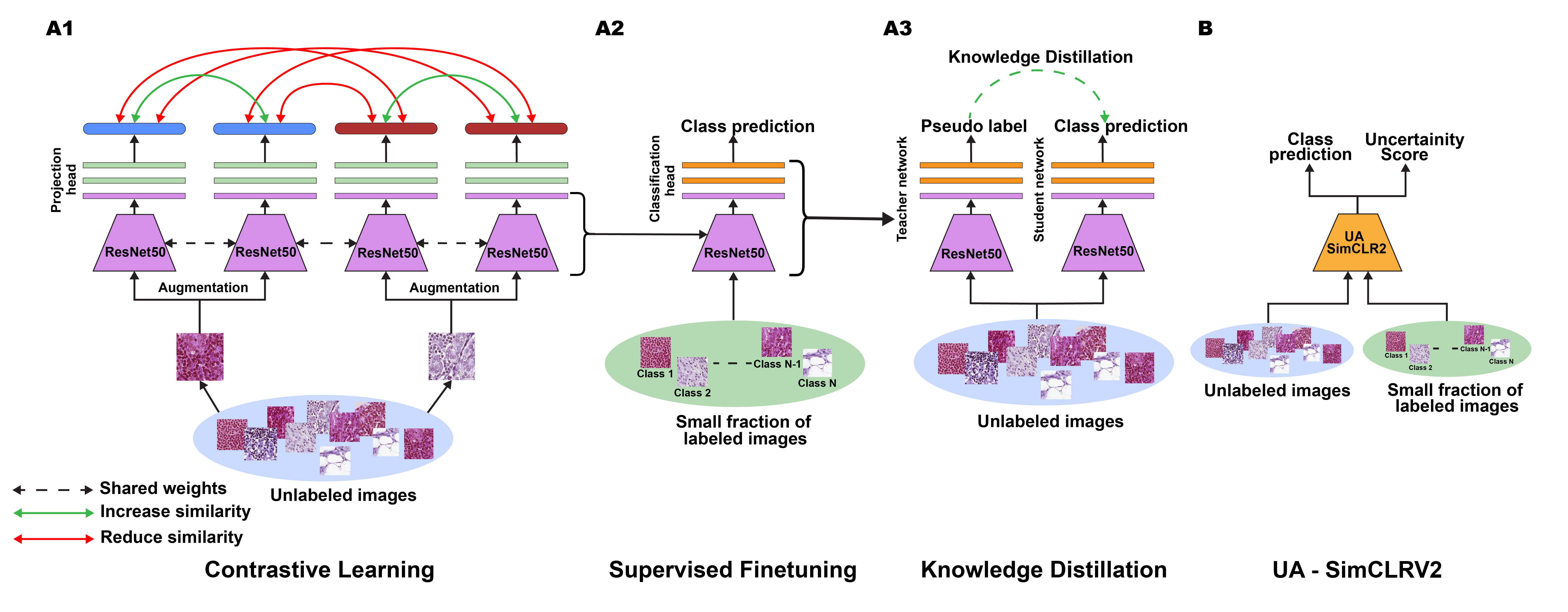}
    \caption{\small The SimCLRv2 framework comprises three steps: (A1) Pre-training employs contrastive learning on unlabelled images. (A2) Supervised fine-tuning adds a classification head to the pre-trained encoder and fine-tunes using labeled images. (A3) Knowledge distillation involves using the fine-tuned model as a teacher network to generate pseudo labels for unlabeled images, then training a student network. (B) The proposed UA-SimCLRv2 model extends this with an additional output for the uncertainty score, enhancing model prediction explainability.}
    \label{fig:simclrv2}
\end{figure*}

\subsection{Datasets}

In this work, we used two datasets, the Patch Camelyon (PCam) dataset \cite{Veeling2018-qh}, and the NCT-CRC-HE-100K (NCT100k) dataset \cite{kather_jakob_nikolas_2018_1214456}. Table~\ref{table: pcamdesc} details the PCam dataset and Table~\ref{table: nctdesc} details the NCT100k dataset.


\begin{table}[htbp]
    \centering
    \small{
    \caption{\small  Description of attributes and characteristics of the PCam dataset extracted from histopathology scans of lymph node sections from CAMELYON16.}
    \label{table: pcamdesc}    
    \begin{tabular}{c c}

        \toprule
            \textbf{Attribute}&\textbf{Description} \\
             
        \midrule
        {Source} & {CAMELYON16 WSI} \\
        \midrule
        {Patch Count} & {327,680} \\
        \midrule
        {Patch Size} & {96x96 pixels (resized to 224x224)} \\
        \midrule
        { Label} & {Binary indicating metastatic tissue presence} \\
        \midrule
        {Data Splits} & {Training (75\%), Validation (12.5\%), Test (12.5\%)} \\
        \bottomrule
    \end{tabular}
    }
\end{table}


\begin{table}[htbp]
    \centering
    \small{
    \caption{\small  Description of attributes and characteristics of the NCT100k dataset, consisting of  H\&E stained histological images  annotated into nine classes. }
    \label{table: nctdesc}    
    \begin{tabular}{c c}

        \toprule
            \textbf{Attribute}&\textbf{Description} \\
             
        \midrule
        {Source} & {Human colorectal cancer } \\
        & {and normal tissues} \\
        \midrule
        {Patch Count} & {100,000} \\
        \midrule
        {Patch Size} & {224x224 pixels} \\
        \midrule
        { Label} & {Adipose (ADI), Background (BACK),   } \\
        & {Debris (DEB), Lymphocytes (LYM), } \\
        & {Mucus (MUC), Smooth Muscle (MUS),} \\
        & {Normal Colon Mucosa (NORM),} \\
        & {Cancer-Associated Stroma (STR), } \\
        & {Colorectal Adenocarcinoma Epithelium (TUM)} \\
        \midrule
        {Data Splits} & {Training (NCT100k),} \\
        & { Validation ( CRC-VAL-HE-7K (CRC7k))} \\
        
        \bottomrule
    \end{tabular}
    }
\end{table}

\subsection{Patch Level Classification}
\label{sec:patchandWSIclassification}

We first utilized SimCLRv2 as a patch-level classifier (see the model architecture in Fig.~\ref{fig:simclrv2}). For both datasets, we benchmarked our model against other models that have been used previously for the same task in digital pathology.  Our experimental framework assessed model performance based on two criteria: the proportion of training set annotations used for fine-tuning and the context of pre-training data (in-domain or out-domain). In-domain refers to using the same dataset for pre-training (contrastive learning), fine-tuning, and knowledge distillation, followed by testing the trained model on the same dataset. Out-domain refers to using a different dataset for the pre-training step. For instance, in the PCam-outdomain setting the NCT100K dataset was used for pre-training.

\subsection{Evaluation Metrics}
\label{sec:evaluation_metrics}

To evaluate the performance of our classification models, we used three primary metrics: Accuracy, F1 Score, and Area Under the ROC Curve (AUC).

\begin{itemize}
    \item \textbf{Accuracy:}  Measures the ratio of correctly classified instances to the total number of instances in the dataset, offering a basic measure of overall model performance. However, accuracy may be less informative when there is a class imbalance.
    \item \textbf{F1 Score:} This balances precision and recall, making it especially useful for evaluating model performance in cases of class imbalance. In this work, we employed the \textit{weighted-average} F1 score, which calculates the F1 score for each class individually and then averages these scores according to the proportion of each class in the dataset. This approach ensures that classes with more instances contribute proportionally to the final score, providing a more balanced evaluation across all classes.
    \item \textbf{Area Under the ROC Curve (AUC):} Measures the area under the ROC curve, which plots the true positive rate against the false positive rate across different threshold values. The AUC score indicates how well the model can distinguish between classes, with values closer to 1.0 representing stronger performance in correctly classifying positive and negative instances. A score of 0.5 implies random performance, while a higher AUC reflects better discriminative ability.
\end{itemize}

These metrics together provide a comprehensive view of the model’s accuracy and its robustness in handling class imbalances.

\subsection{UA SimCLRv2}
\label{sec:UQ}

We next introduced uncertainty awareness \cite{sensoy2018evidential} to the SimCLRv2 framework. Our uncertainty aware SimCLRv2 is termed UA-SimCLRv2. The primary objective of UA-SimCLRv2 is to enhance the interpretability of the model's predictions in the context of digital pathology. This is achieved by incorporating the theory of uncertainty estimation, which serves as the basis for uncertainty awareness in UA SimCLRv2.

In \cite{sensoy2018evidential}, the uncertainty estimation is approached from  Dempster–Shafer theory of evidence (DST) perspective \cite{dempster2008upper} assigning belief masses to subsets of a frame of discernment, which denotes the set of exclusive possible states. Subjective logic formalizes DST’s notion of belief assignments over a frame of discernment as a Dirichlet distribution. Term evidence is a measure of the amount of support collected from data in favor of a sample to be classified into a certain class. Through model training evidence $e_k$ ($k=1,2,\ldots,K$) are collected and belief masses $b_k$ ($k=1,2,\ldots,K$) are assigned to each class based on the evidence collected and the remaining are marked as uncertainty $u$. For $K$ mutually exclusive classes,

\begin{equation}\label{eq:uncer plus bm}
    u + \sum_{k=1}^{K}b_k = 1.
\end{equation}
Here $u\geq0$ and $b_k\geq0$, and they are calculated by,
\begin{equation}\label{eq:uncer and bm}
    b_k=\frac{e_k}{S}  \quad \text{and,} \quad  u=\frac{K}{S}, \quad \text{where} \quad S=\sum_{i=1}^{K}e_i + 1.
\end{equation}
Observe that when there is no evidence, the belief for each class is zero and the uncertainty is one. A belief mass assignment, i.e., subjective opinion, corresponds to a Dirichlet distribution with parameters $\alpha_k=e_k + 1$. A Dirichlet distribution parameterized over evidence represents the density
of each such probability assignment; hence it models second-order probabilities and uncertainty \cite{josang2016generalising}. It is characterized by $K$ parameters $\alpha=[\alpha_1,\alpha_2,\ldots,\alpha_K]$ and is given as
\label{eq:dirichlet}
\[
{
    D(p\|\alpha) =
    \begin{cases}
    \frac{1}{B(\alpha)}\prod_{i=1}^{K}p_i^{\alpha_i - 1}, & \text{if } p \in S_K \\
    0, & \text{otherwise,}
\end{cases}
}  
\]
where $S_K = \left\{p\| \sum_{i=1}^{K}p_i =1 \ \text{and} \  0\leq p_1,\ldots,p_k \leq 1 \right\}$ and $B(\alpha)$  is the $K$-dimensional multinomial beta function \cite{kotz2000continuous}.

Model training follows the classical neural network architecture with a softmax layer replaced with ReLU activation layer to ascertain non-negative output, which is taken as the evidence vector for the predicted Dirichlet distribution. For network parameters $\theta$, let $f(x_i\|\theta)$ be the evidence vector predicted by the network for the classification. Corresponding Dirichlet distribution's parameters $\alpha_i=f(x_i\|\theta) + 1$ are calculated and their means $(\frac{\alpha_i}{S})$ are considered as the class probabilities. Let $y_i$ be one hot vector encoding the ground-truth class label of a sample $x_i$. Treating $D(p_i\|\alpha_i)$ as a prior on the sum of squares loss $\|y_i - p_i\|_2^2$, we obtain the loss function 
\begin{align}\label{eq:mseloss}
L_i(\theta)=\int\|y_i - p_i\|_2^2\frac{1}{B(\alpha_i)}\prod_{i=1}^{K}p_{ij}^{\alpha_{ij} - 1}dp_i.
\end{align}

By decomposing the first and second moments, minimization of both the prediction error and the variance of the Dirichlet experiment for each sample is achieved by the above loss function. Further some evidence collected might strengthen the belief for multiple classes. To avoid situations where evidence with more ambiguity assigns more belief to incorrect class,  Kullback-Leibler (KL) divergence term is appended to the loss function. Following is the total loss used for UA fine-tuning.

\begin{align}\label{eq:loss}
L(\theta)=&\sum_{i=1}^NL_i(\theta) \notag \\ &\hspace{-5mm}+\lambda_t\sum_{i=1}^NKL[D(p_i\|\Tilde{\alpha_i})\|D(p_i\|<1,\ldots,1>)]
\end{align}
where $\lambda_t=\min(1,t/10) \in [0,1]$ is the annealing coefficient, $t$ is the index of the current training epoch, $D(p_i\|<1,\ldots,1>)$ is the uniform Dirichlet distribution, and $\Tilde{\alpha_i}=y_i+(1-y_i)*\alpha_i$ is the Dirichlet parameters after removal of the non-misleading evidence from predicted parameters $\alpha_i$ for sample $i$. The KL divergence term in the loss can be calculated as

\begin{multline*}\label{eq:KL}
KL[D(p_i\|\Tilde{\alpha_i})||D(p_i\|1)] \\
= \log\bigg(\frac{\Gamma(\sum_{k=1}^K\Tilde{\alpha_{ik}})}{\Gamma(K)\prod_{k=1}^K\Gamma(\Tilde{\alpha_{ik}})}\bigg) \\
+ \sum_{k=1}^K(\Tilde{\alpha_{ik}}-1)\bigg[\psi(\Tilde{\alpha_{ik}}) - \psi\bigg(\sum_{j=1}^K\Tilde{\alpha_{ij}}\bigg)\bigg] 
\end{multline*}
where 1 represents the parameter vector of $K$ ones, $\Gamma(\cdot)$ is the gamma function, and $\psi(\cdot)$ is the digamma function. By gradually increasing the effect of the KL divergence in the loss through the annealing coefficient, the neural network is allowed to explore the parameter space and avoid premature convergence to the uniform distribution for the misclassified samples, which may be correctly classified in future epochs.

\subsection{Uncertainty-aware Active learning (UA-AL)}
\label{sec:uafinetuning}

\begin{figure*}[htbp]
    \centering
    \includegraphics[width=\linewidth]{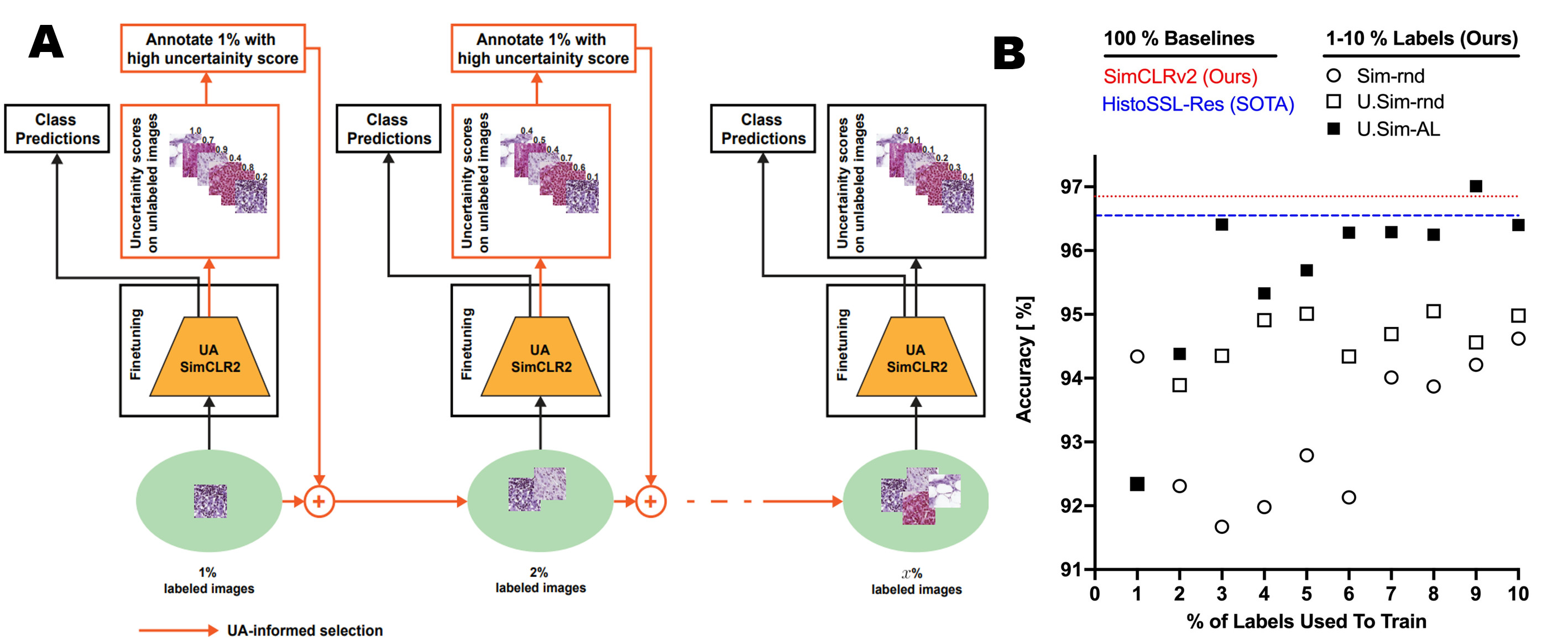}
    \caption{\small (A) UA-AL framework (B) Outdomain training of UA-AL outperforms random label selection, achieving comparable results to the SOTA with just 3\% of labels, and surpassing the SOTA with 9\% of labels. }
    \label{fig:fig2}
\end{figure*}


Last, we leveraged uncertainty scores in UA-SimCLRv2, as the querying strategy for AL. We demonstrated our AL for subtyping of NCT100k dataset. Starting with a pre-trained model, we first labeled 1\% of images randomly and fine-tuned the model. Subsequently, we iteratively queried the top 1\% uncertain images and added them to the training set with expert-annotated labels. This process continued until 10\% of the labels were used in training. As in Fig. \ref{fig:fig2}.A, We compared UA-AL with both regular SimCLRv2 and UA-SimCLRv2 models with random sampling for labeling at each iteration, assessing model performance on the test dataset. As discussed in section \ref{subsec:al}, because the related work showed comparable or slightly lower accuracy than the random sampling strategy, we conducted the comparison directly against the random sampling method.

%% file: Results.tex
\section{Results}
\label{sec:experiments}

In Section~\ref{sec:Patch level Classification}, we evaluate various SSL frameworks for patch-level classification to identify the most effective one. We then incorporate uncertainty awareness into the chosen SSL framework. Then, we visualize and analyze the suitability of the selected uncertainty estimation method as a querying strategy in AL. Finally, in Section~\ref{sec:Uncertainty-aware Label Selection}, we discuss the effectiveness of uncertainty awareness in AL and observe that we achieve SOTA results in patch-level classification using only 2\% of in-domain labels.

\subsection{Patch level Classification}
\label{sec:Patch level Classification}

\subsubsection{Binary Class Classification}

\begin{table*}[htbp]
    \centering
    \captionsetup{justification=centering}
    \small{
    \caption{\small Binary classification results for PCam dataset for a variety of models.  Results marked by $*$ are quoted from \cite{luo2022self}.}
    \label{table: pcam2}    
    \begin{tabular}{l  l  l ccc ccc}

        \toprule
          & &  &
            \multicolumn{3}{c}{Regular Model}  &
            \multicolumn{3}{c}{Uncertainty-aware Model} \\
            \multicolumn{2}{c}{\textbf{Labels}}&\textbf{Model} & Acc (\%) & F1 (\%) & AUC (\%) & Acc (\%) & F1 (\%) & AUC (\%) \\
             
        \midrule
        &Indomain&TransPath$^*$ \cite{wang2021transpath}   & 81.20 & 81.00 & 91.70 & - & - & -\\
        &Indomain&Mocov3$^*$ \cite{chen2021mocov3}  & 86.30 & 86.20 & 95.00 & - & - & -\\
        &Indomain&DINO$^*$ \cite{caron2021emerging} & 85.80 & 85.60 & 95.70 & - & - & -\\
        100\%&Indomain&SD-MAE$^*$ \cite{luo2022self} & 88.20 & 87.80 & 96.20 & - & - & -\\
        &Indomain&MAE \cite{mae}  & 88.41 & 86.23 & 95.81 &  - & - & -\\
        &Indomain&SimCLRv1 \cite{chen2020simple}  & 83.21 & 84.40 & 88.67 & - & - & -\\
        &Indomain&\textbf{SimCLRv2} \cite{chen2020big}   
 & 90.57 & 90.20 & 96.47 &  90.29 & 89.95& 96.49\\
        &Outdomain&\textbf{SimCLRv2} \cite{chen2020big}  & 89.30& 88.97 & 96.58 &  \textbf{91.30} & \textbf{91.09} & \textbf{96.83} \\
        \midrule
        &Indomain&MAE \cite{mae}  & 86.10 & 84.45 & 94.81 &  - & - & -\\
        10\%&Indomain&SimCLRv1 \cite{chen2020simple}   & 88.67 & 81.52 & 83.45 & - & - & - \\
        &Indomain&\textbf{SimCLRv2} \cite{chen2020big}   &  89.73 & 89.07 & 96.19  & 88.27& 88.94 & 94.69 \\
        &Outdomain&\textbf{SimCLRv2} \cite{chen2020big}  & 89.60 & 88.84& 96.73 &  \textbf{90.41} & \textbf{89.97} & \textbf{96.87} \\
        \midrule
        &Indomain&MAE \cite{mae}  & 85.81 & 86.10 & 94.45 &  - & - & -\\
        1\%&Indomain&SimCLRv1 \cite{chen2020simple}  & 87.77 & 88.67 & 81.52 & - & - & - \\
        &Indomain&\textbf{SimCLRv2} \cite{chen2020big}    &  \textbf{90.27}& \textbf{89.99} & \textbf{95.34}  & 88.96& 88.54 & 94.24 \\
        &Outdomain&\textbf{SimCLRv2} \cite{chen2020big}   & 89.21 & 88.88 & 95.57 &  87.43 & 86.96& 92.33\\
        \bottomrule
    \end{tabular}
    }
\end{table*}

Table~\ref{table: pcam2} show the accuracy, F1 score, and AUC score for the  PCam binary classification. To establish a baseline, we first fine-tuned our models with all training labels (i.e., the 100\% setting). Here, our models outperformed the SOTA approach, i.e., MAE \cite{mae}. In-domain pre-trained SimCLRv2, performed best with 2.16\% increase in accuracy compared to the SOTA and UA-SimCLRv2 performed even better.  Next, we fine-tuned our models on 10\%  training labels. 10\%-fine-tuned models performed slightly worse than the 100\% baseline. Nevertheless, the 10\%-fine-tuned SimCLRv2 and  UA-SimCLRv2 still performed on par with or better than the SOTA. Then, we fine-tuned our models on 1\% training labels. Interestingly the SimCLRv2 and UA-SimCLRv2 models still performed comparable to the SOTA (see the 1\% setting on Table~\ref{table: pcam2}). However, at the 1\% setting UA-SimCLRv2 consistently underperformed compared to SimCLRv2, perhaps due to the limited evidence available for uncertainty awareness. Rows in bold highlight the best results within their respective sections.

\subsubsection{Multi-Class Classification}

\begin{table*}[htbp]
    \centering
    \captionsetup{justification=centering}
    \small{
    \caption{\small Multi-class classification results for NCT100k dataset for a variety of models. Results marked by $*$ are quoted from \cite{luo2022self}; Results marked by $**$ are quoted from \cite{jin2022histossl}.}
    \label{table: nct2}    
    \begin{tabular}{l  l  l  cc cc}

        \toprule
          & &  &
            \multicolumn{2}{c}{Regular Model}  &
            \multicolumn{2}{c}{Uncertainty-aware Model} \\
            \multicolumn{2}{c}{\textbf{Labels}}&\textbf{Model} & Acc (\%) & F1 (\%) & Acc (\%) & F1 (\%)  \\
             
        \midrule
        &Indomain&TransPath$^*$ \cite{wang2021transpath}  & 92.80 & 89.90  & - & - \\
        &Indomain&Mocov3$^*$ \cite{chen2021mocov3}  & 94.40 & 92.60  & - & -\\
        &Indomain&DINO$^*$ \cite{caron2021emerging} & 94.40 & 91.60 & - & - \\
        &Indomain  & BYOL$^{**}$ \cite{byol}  & 93.93 & -   & - & -\\
        & Indomain &HistoSSL-Res$^{**}$ \cite{jin2022histossl}  & 96.55 & - & - & - \\
        100\%& Indomain &HistoSSL-ViT$^{**}$ \cite{jin2022histossl} & 96.18 & - &  - & - \\  
        &Indomain &SD-MAE$^*$ \cite{luo2022self} & 95.30 & 93.50  & - & - \\
        &Indomain&MAE \cite{mae}  & 94.70 & 94.20   & - & -\\
        &Indomain&SimCLRv1 \cite{chen2020simple}  & 92.10 & 92.20 & -   & - \\
        &Indomain&\textbf{SimCLRv2} \cite{chen2020big}  & 96.28  & 96.25  & 96.44 & 96.39\\
        &Outdomain&\textbf{SimCLRv2} \cite{chen2020big}   & \textbf{96.85}  & \textbf{96.82}   & 95.88 & 95.82\\
        \midrule
        10\%&Indomain&\textbf{SimCLRv2} \cite{chen2020big}  
 & \textbf{96.28}  & \textbf{96.25} & 95.82 & 95.73 \\
        &Outdomain&\textbf{SimCLRv2} \cite{chen2020big}  & 94.62  & 94.56  & 94.98 & 94.87\\
        \midrule
        1\%&Indomain&\textbf{SimCLRv2} \cite{chen2020big}   & 94.27  & 94.12& 91.70 & 91.65 \\
        &Outdomain&\textbf{SimCLRv2} \cite{chen2020big}  
 & \textbf{94.34}  & \textbf{94.23} & 92.34 & 92.85 \\
        
        \bottomrule
    \end{tabular}
    }
\end{table*}

Table~\ref{table: nct2} shows multi-class classification results for the NCT100k dataset. Similar to the binary case, we experimented at 100\%, 10\% and 1\% fine-tuning settings. First, at the 100\% setting our SimCLRv2 and UA-SimCLRv2 performed on par with the SOTA. Interestingly, out-domain pre-trained SimCLRv2 was the best-performing model and surpassed the SOTA by a small margin. But at the 1\% setting, we observed a degradation of performance by a few percentage points. We would further explore the impact of in-domain and out-domain setting in future work.

Tables ~\ref{table: pcam2} and ~\ref{table: nct2} demonstrate that SimCLRv2 stands out as the superior SSL framework when compared to current SOTA models. Furthermore, the incorporation of uncertainty awareness into SimCLRv2 not only maintains its high accuracy but also enhances the interpretability of its predictions.

\subsubsection{Visualizing Uncertainty estimation}
\label{sec:Visualizing Uncertainty estimation}
\begin{figure*}[htbp]
    \centering
    \includegraphics[width=\linewidth]{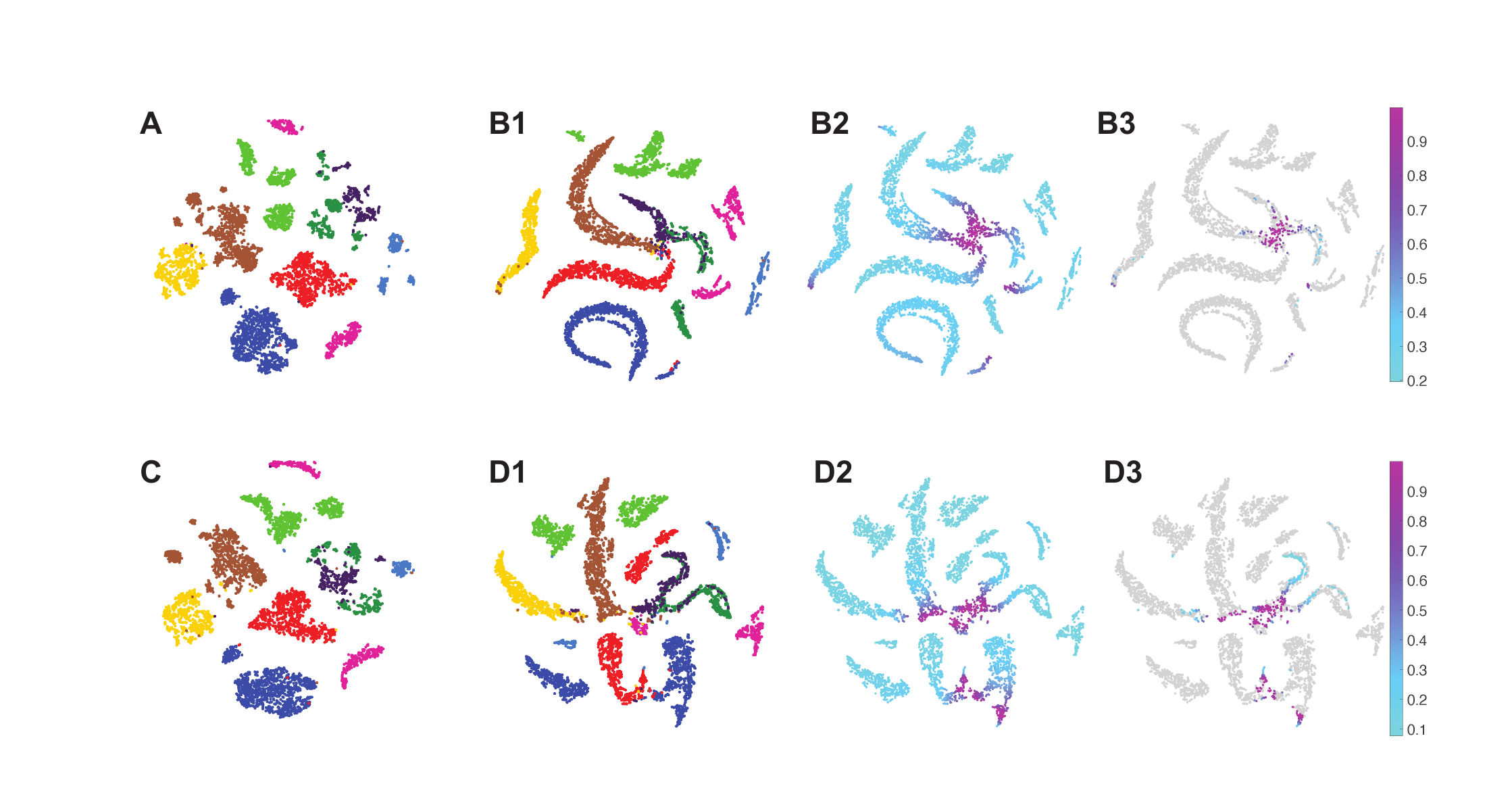}
   \vspace{-8mm}
    \caption{\small T-SNE plot (A) for SimCLRv2 trained in distribution with 100\% annotations (B1) for UA-SimCLRv2 trained in distribution with 100\% annotations  (B2) color coded with the uncertainty values (Note that mixed cluster regions show high uncertainty) (B3) where only the Incorrect predictions are color coded. Note that most incorrect predictions show high uncertainty.  C, D1, D2, D3 Corresponding versions of ‘A, B1, B2, B3’ with 1\% of annotations. Note that in ‘D3’ there are more incorrect predictions with low uncertainty values than in ‘B3’.}
    \label{fig:fig3}
\end{figure*}

\begin{table*}[htbp]
   \centering
   \captionsetup{justification=centering}
    \small{
    \caption{\small Results of UA-AL and random sampling of labels on the NCT100K dataset. rnd - Random Labeling, Sim - SimCLRv2, U.S - UA-SimCLRv2}
    \label{table: uatraining}    
    \begin{tabular}{l | cc | cc | cc | cc | cc | cc}

        \toprule
        & \multicolumn{6}{c}{Indomain} & \multicolumn{6}{c}{Outdomain} \\
        
           \textbf{Lab.} & \multicolumn{2}{c}{Sim-rnd} &\multicolumn{2}{c}{U.Sim-rnd } &\multicolumn{2}{c}{U.Sim-AL } & \multicolumn{2}{c}{Sim-rnd} &\multicolumn{2}{c}{U.Sim-rnd } &\multicolumn{2}{c}{U.Sim-AL }  \\
           &Acc-\% & F1-\% & Acc-\% & F1-\% &Acc-\% & F1-\% & Acc-\% & F1-\%  &Acc-\% & F1-\%  & Acc-\%  & F1-\%   \\
             
        \midrule
        \textbf{1\%}  & 94.27 & 94.12 & 91.70 & 91.65 & 91.70 & 91.65 & 94.34 & 94.22  & 92.34 & 92.25  & 92.34 & 92.25  \\
       \textbf{2\%}  &93.57 & 93.46  & 94.59  & 94.13    & 96.26 & 96.15 & 92.31 & 92.23   & 93.89 & 93.58   & 94.38 & 94.34  \\
        \textbf{3\%}  &  92.01 & 91.87  & 93.23 & 92.95 & 96.29 & 96.23 & 91.67 & 91.65   & 94.35  & 94.32  & 96.41 & 96.31  \\
        \textbf{4\%}  & 91.22 & 91.20  & 95.30 & 95.18  & 95.35 & 95.35 & 91.98 & 91.92 & 94.91 & 94.90  & 95.33 & 95.21  \\
        \textbf{5\%}  & 91.69 & 91.69 & 94.56 & 94.49 & 96.03 & 96.02 & 92.79 & 92.45  & 95.01 & 94.93  & 95.69 & 95.68  \\
        \textbf{6\%}  & 91.68 & 91.65 & 94.06 & 93.94 & 95.93 & 95.91 & 92.13 & 92.11  & 94.34 & 94.21   & 96.28 & 96.25  \\ 
        \textbf{7\%}  & 92.48 & 92.42 & 95.12  & 94.91  & 95.76 & 95.76 & 94.01  & 93.97  & 94.69&94.56   & 96.29 & 96.25  \\
        \textbf{8\%}  & 92.08 & 92.05  & 95.01 & 94.89 & 96.50 & 96.45 & 93.87 & 93.46  & 95.05 & 94.98  & 96.25 & 96.12  \\
        \textbf{9\%}  & 94.62 & 94.54 & 96.32  & 96.28 & 96.51 & 96.42 & 94.21 & 94.12  & 94.56 & 94.32  & \textbf{97.01} & \textbf{96.90}  \\
        \textbf{10\%}  & 96.28 & 96.25 & 95.82 & 95.73 & 96.51 & 96.49 & 94.62 & 94.56  & 94.98 & 94.87  & 96.40 & 96.33  \\
        
        \bottomrule
    \end{tabular}
    }
\end{table*}
Our t-SNE analysis in Fig.~\ref{fig:fig3} initially showed that compared to SimCLRv2, UA-SimCLRv2's T-SNE maps demonstrate improved interpretability, characterized by better cluster border refinement and organization of points based on prediction uncertainty (refer Fig.~\ref{fig:fig3}.B2). Furthermore, it was observed that more interpretable predictions (i.e., incorrect predictions with higher uncertainty scores) were attainable when a larger number of labels were available. This highlights the effectiveness of uncertainty estimation in discerning prediction reliability. 

\begin{figure*}[htbp]
    \centering
    \includegraphics[width=\linewidth]{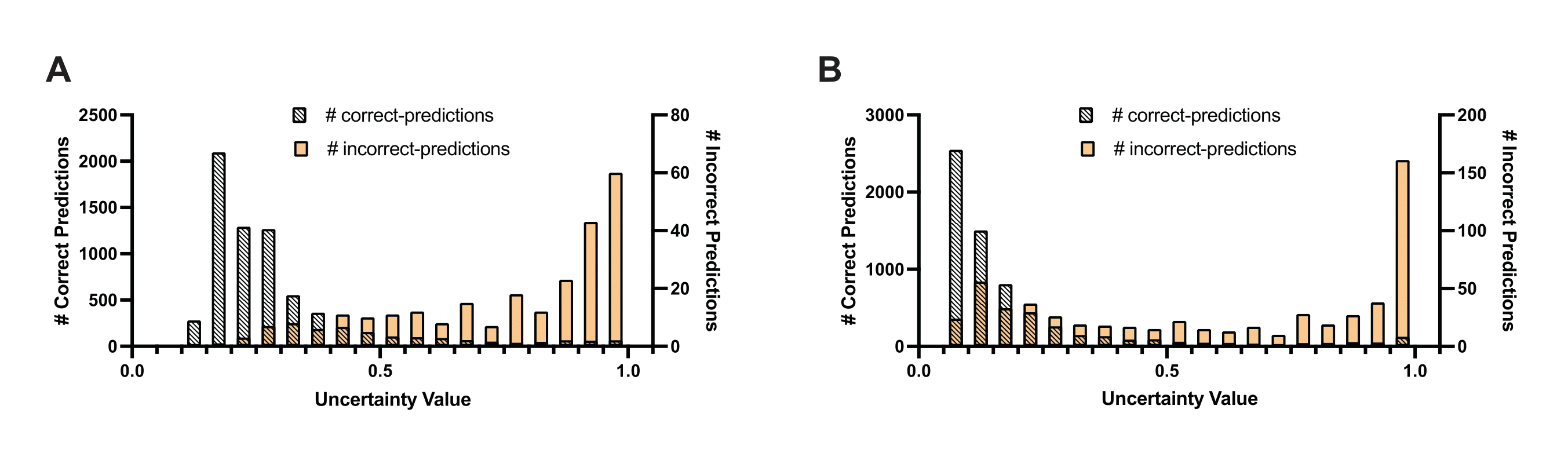}
    \caption{\small Histograms for (A)  100\% annotations (B) 1\% annotations demonstrating the tight coupling between model predictions accuracy and uncertainty awareness. }
    \label{fig:fig4}
\end{figure*}

We also plotted the histograms of uncertainty values of correct and incorrect predictions (see Fig.~\ref{fig:fig4}). Note that incorrect prediction histograms correspond to ‘B3’ and ‘D3’ of Fig.~\ref{fig:fig3} . 
In both the 100\% and 1\% settings, correct predictions exhibited a left-skewed distribution, while incorrect predictions displayed a right-skewed distribution. This observation indicates that the majority of incorrect predictions correlate with high uncertainty, whereas correct predictions tend to exhibit low uncertainty. This alignment underscores the effectiveness of the uncertainty estimation method in enhancing the interpretability of model predictions. The insight from Fig.~\ref{fig:fig3} \& Fig.~\ref{fig:fig4} enabled us to develop a querying strategy for UA-AL.

\subsection{Uncertainty-aware Label Selection}
\label{sec:Uncertainty-aware Label Selection}

Table~\ref{table: uatraining} showcases the accuracy and F1 score outcomes from training the UA-SimCLRv2 model, utilizing an uncertainty-aware label selection approach. This is set against the backdrop of the SimCLRv2 model's performance, as well as the UA-SimCLRv2 model when trained with a random image selection for labeling. In scenarios where training is conducted for indomain, we observed that the accuracy and F1 scores rapidly approached the baseline set by the 100\% label setting with just 2\% of labels. However, beyond this point, performance gains began to plateau. Notably, the UA-training method consistently outperformed models fine-tuned with randomly selected labels. The peak performance was recorded in an out-domain pre-training context, where it achieved superior results (refer Fig.~\ref{fig:fig2}.B) with only 9\% of labels using uncertainty-aware labeling. This model not only surpassed the current state-of-the-art, HistoSSL-Res~\cite{jin2022histossl}, but also outperformed the baseline model trained with 100\% labels (as detailed in Table.~\ref{table: nct2}).

The implications of these findings are threefold: firstly, UA-SimCLRv2 emerges as the foremost patch classifier on the NCT100k benchmarks. Secondly, even with randomly selected labels, UA-SimCLRv2 outperforms SimCLRv2 as label quantity increases. Lastly, employing uncertainty-aware label selection consistently leads to higher accuracy compared to random selection methods.

%% file: Conclusion.tex
\section{Conclusion}
\label{sec:conclusion}

Our research represents a significant advancement in cancer subtyping for digital pathology, by integrating uncertainty awareness into SSL and AL frameworks. The UA-SimCLRv2 model offers superior interpretability in model predictions and performance, surpassing SOTA approaches with minimal labeled data. By strategically querying uncertain samples for annotation, our framework not only reduces annotation burdens but also enhances model precision and efficiency. These findings underscore the importance of incorporating uncertainty awareness into the learning process, particularly in critical domains like digital pathology.  

With UA-SimCLRv2 established as the leading classifier on digital pathology benchmark datasets, our research sets a new standard in cancer subtyping in histopathology. This work can further be extended to whole slide image classification by using our fine-tuned encoder as the backbone to MIL approach. In essence, our work transforms digital pathology image analysis, introducing a new era of precision and efficiency led by the UA-AL with UA-SimCLRv2.

\newpage
\newpage

%% file: main.bbl
\begin{thebibliography}{10}\itemsep=-1pt

\bibitem{caron2021emerging}
Mathilde Caron, Hugo Touvron, Ishan Misra, Herv{\'e} J{\'e}gou, Julien Mairal, Piotr Bojanowski, and Armand Joulin.
\newblock Emerging properties in self-supervised vision transformers.
\newblock In {\em Proceedings of the IEEE/CVF international conference on computer vision}, pages 9650--9660, 2021.

\bibitem{carse2019active}
Jacob Carse and Stephen McKenna.
\newblock Active learning for patch-based digital pathology using convolutional neural networks to reduce annotation costs.
\newblock In {\em Digital Pathology: 15th European Congress, ECDP 2019, Warwick, UK, April 10--13, 2019, Proceedings 15}, pages 20--27. Springer, 2019.

\bibitem{chen2020simple}
Ting Chen, Simon Kornblith, Mohammad Norouzi, and Geoffrey Hinton.
\newblock A simple framework for contrastive learning of visual representations.
\newblock In {\em International conference on machine learning}, pages 1597--1607. PMLR, 2020.

\bibitem{chen2020big}
Ting Chen, Simon Kornblith, Kevin Swersky, Mohammad Norouzi, and Geoffrey~E Hinton.
\newblock Big self-supervised models are strong semi-supervised learners.
\newblock {\em Advances in neural information processing systems}, 33:22243--22255, 2020.

\bibitem{sim}
Xinlei Chen and Kaiming He.
\newblock Exploring simple siamese representation learning.
\newblock In {\em Proceedings of the IEEE/CVF conference on computer vision and pattern recognition}, pages 15750--15758, 2021.

\bibitem{chen2021mocov3}
Xinlei Chen*, Saining Xie*, and Kaiming He.
\newblock An empirical study of training self-supervised vision transformers.
\newblock {\em arXiv preprint arXiv:2104.02057}, 2021.

\bibitem{SelfSupervisedHisto}
Ozan Ciga, Tony Xu, and Anne~Louise Martel.
\newblock Self supervised contrastive learning for digital histopathology.
\newblock {\em Machine Learning with Applications}, 7:100198, 2022.

\bibitem{dempster2008upper}
Arthur~P Dempster et~al.
\newblock Upper and lower probabilities induced by a multivalued mapping.
\newblock {\em Classic works of the Dempster-Shafer theory of belief functions}, 219(2):57--72, 2008.

\bibitem{dolezal2022uncertainty}
James~M Dolezal, Andrew Srisuwananukorn, Dmitry Karpeyev, Siddhi Ramesh, Sara Kochanny, Brittany Cody, Aaron~S Mansfield, Sagar Rakshit, Radhika Bansal, Melanie~C Bois, et~al.
\newblock Uncertainty-informed deep learning models enable high-confidence predictions for digital histopathology.
\newblock {\em Nature communications}, 13(1):6572, 2022.

\bibitem{du2018breast}
Baolin Du, Qi Qi, Han Zheng, Yue Huang, and Xinghao Ding.
\newblock Breast cancer histopathological image classification via deep active learning and confidence boosting.
\newblock In {\em International Conference on Artificial Neural Networks}, pages 109--116. Springer, 2018.

\bibitem{gal2016dropout}
Yarin Gal and Zoubin Ghahramani.
\newblock Dropout as a bayesian approximation: Representing model uncertainty in deep learning.
\newblock In {\em international conference on machine learning}, pages 1050--1059. PMLR, 2016.

\bibitem{obow}
Spyros Gidaris, Andrei Bursuc, Nikos Komodakis, Patrick P{\'e}rez, and Matthieu Cord.
\newblock Learning representations by predicting bags of visual words.
\newblock In {\em Proceedings of the IEEE/CVF Conference on Computer Vision and Pattern Recognition}, pages 6928--6938, 2020.

\bibitem{byol}
Jean-Bastien Grill, Florian Strub, Florent Altch{\'e}, Corentin Tallec, Pierre Richemond, Elena Buchatskaya, Carl Doersch, Bernardo Avila~Pires, Zhaohan Guo, Mohammad Gheshlaghi~Azar, et~al.
\newblock Bootstrap your own latent-a new approach to self-supervised learning.
\newblock {\em Advances in neural information processing systems}, 33:21271--21284, 2020.

\bibitem{hang2022reliability}
Wenlong Hang, Yecheng Huang, Shuang Liang, Baiying Lei, Kup-Sze Choi, and Jing Qin.
\newblock Reliability-aware contrastive self-ensembling for semi-supervised medical image classification.
\newblock In {\em International Conference on Medical Image Computing and Computer-Assisted Intervention}, pages 754--763. Springer, 2022.

\bibitem{mae}
Kaiming He, Xinlei Chen, Saining Xie, Yanghao Li, Piotr Doll{\'a}r, and Ross Girshick.
\newblock Masked autoencoders are scalable vision learners.
\newblock In {\em Proceedings of the IEEE/CVF conference on computer vision and pattern recognition}, pages 16000--16009, 2022.

\bibitem{hinton2015distilling}
Geoffrey Hinton, Oriol Vinyals, and Jeff Dean.
\newblock Distilling the knowledge in a neural network.
\newblock {\em arXiv preprint arXiv:1503.02531}, 2015.

\bibitem{jin2021reducing}
Xu Jin, Hong An, Jue Wang, Ke Wen, and Zheng Wu.
\newblock Reducing the annotation cost of whole slide histology images using active learning.
\newblock In {\em Proceedings of the 2021 3rd International Conference on Image Processing and Machine Vision}, pages 47--52, 2021.

\bibitem{jin2022histossl}
Xu Jin, Teng Huang, Ke Wen, Mengxian Chi, and Hong An.
\newblock Histossl: Self-supervised representation learning for classifying histopathology images.
\newblock {\em Mathematics}, 11(1):110, 2022.

\bibitem{josang2016generalising}
Audun J{\o}sang.
\newblock Generalising bayes' theorem in subjective logic.
\newblock In {\em MFI}, pages 462--469, 2016.

\bibitem{kather_jakob_nikolas_2018_1214456}
Jakob~Nikolas Kather, Niels Halama, and Alexander Marx.
\newblock 100,000 histological images of human colorectal cancer and healthy tissue.
\newblock {\em Zenodo10}, 5281, 2018.

\bibitem{kotz2000continuous}
S Kotz, N Balakrishnan, and NL Johnson.
\newblock Continuous multivariate distributions--vol. 1, john wiley \& sons, new york, 2000.

\bibitem{lakshminarayanan2017simple}
Balaji Lakshminarayanan, Alexander Pritzel, and Charles Blundell.
\newblock Simple and scalable predictive uncertainty estimation using deep ensembles.
\newblock {\em Advances in neural information processing systems}, 30, 2017.

\bibitem{luo2022self}
Yang Luo, Zhineng Chen, and Xieping Gao.
\newblock Self-distillation augmented masked autoencoders for histopathological image classification.
\newblock {\em arXiv preprint arXiv:2203.16983}, 2022.

\bibitem{Misra2019SelfSupervisedLO}
Ishan Misra and Laurens van~der Maaten.
\newblock Self-supervised learning of pretext-invariant representations.
\newblock In {\em Proceedings of the IEEE/CVF conference on computer vision and pattern recognition}, pages 6707--6717, 2020.

\bibitem{sensoy2018evidential}
Murat Sensoy, Lance Kaplan, and Melih Kandemir.
\newblock Evidential deep learning to quantify classification uncertainty.
\newblock {\em Advances in neural information processing systems}, 31, 2018.

\bibitem{seth2023fusdom}
Ashish Seth, Sreyan Ghosh, S Umesh, and Dinesh Manocha.
\newblock Fusdom: Combining in-domain and out-of-domain knowledge for continuous self-supervised learning.
\newblock {\em arXiv preprint arXiv:2312.13026}, 2023.

\bibitem{settles2011theories}
Burr Settles.
\newblock From theories to queries: Active learning in practice.
\newblock In {\em Active learning and experimental design workshop in conjunction with AISTATS 2010}, pages 1--18. JMLR Workshop and Conference Proceedings, 2011.

\bibitem{what}
Yonglong Tian, Chen Sun, Ben Poole, Dilip Krishnan, Cordelia Schmid, and Phillip Isola.
\newblock What makes for good views for contrastive learning?
\newblock {\em Advances in neural information processing systems}, 33:6827--6839, 2020.

\bibitem{Veeling2018-qh}
Bastiaan~S Veeling, Jasper Linmans, Jim Winkens, Taco Cohen, and Max Welling.
\newblock Rotation equivariant cnns for digital pathology.
\newblock In {\em Medical Image Computing and Computer Assisted Intervention--MICCAI 2018: 21st International Conference, Granada, Spain, September 16-20, 2018, Proceedings, Part II 11}, pages 210--218. Springer, 2018.

\bibitem{wang2013fast}
Sida Wang and Christopher Manning.
\newblock Fast dropout training.
\newblock In {\em international conference on machine learning}, pages 118--126. PMLR, 2013.

\bibitem{wang2021transpath}
Xiyue Wang, Sen Yang, Jun Zhang, Minghui Wang, Jing Zhang, Junzhou Huang, Wei Yang, and Xiao Han.
\newblock Transpath: Transformer-based self-supervised learning for histopathological image classification.
\newblock In {\em Medical Image Computing and Computer Assisted Intervention--MICCAI 2021: 24th International Conference, Strasbourg, France, September 27--October 1, 2021, Proceedings, Part VIII 24}, pages 186--195. Springer, 2021.

\bibitem{wenzel2020hyperparameter}
Florian Wenzel, Jasper Snoek, Dustin Tran, and Rodolphe Jenatton.
\newblock Hyperparameter ensembles for robustness and uncertainty quantification.
\newblock {\em Advances in Neural Information Processing Systems}, 33:6514--6527, 2020.

\bibitem{barlow}
Jure Zbontar, Li Jing, Ishan Misra, Yann LeCun, and St{\'e}phane Deny.
\newblock Barlow twins: Self-supervised learning via redundancy reduction.
\newblock In {\em International Conference on Machine Learning}, pages 12310--12320. PMLR, 2021.

\end{thebibliography}
